\pdfoutput=1
\documentclass[10pt,twocolumn,letterpaper]{article}
\usepackage[pagenumbers]{cvpr}  

\usepackage{newtxtext,newtxmath}  
\usepackage{float}      
\usepackage{microtype}   
\definecolor{cvprblue}{rgb}{0.21,0.49,0.74}
\usepackage[pagebackref,breaklinks,colorlinks,allcolors=cvprblue]{hyperref}
\hypersetup{pdftitle={What Do Audio-Visual Synchronization Metrics Actually Measure?},pdfauthor={Jai Kumar Sharma, Peeyush Tapadiya},pdfsubject={Reliability audit of audio-visual synchronization metrics (Gen4AVC @ ECCV 2026)},pdfkeywords={audio-visual synchronization, evaluation metrics, meta-evaluation, audio-visual generation, AV-Align, Synchformer, ImageBind, JavisScore, PEAVS}}
\title{What Do Audio-Visual Synchronization Metrics Actually Measure?\thanks{Accepted at the ECCV 2026 Workshop on Generative AI for Audio-Visual Content Creation (Gen4AVC), poster presentation (non-archival). Project page: \protect\url{https://jaishrm07.github.io/avsync-reliability-card/}}}
\author{Jai Kumar Sharma\\
Virginia Tech\\
{\tt\small jaisharma@vt.edu}
\and
Peeyush Tapadiya\\
Accenture\\
{\tt\small peeyush.tapadiya@accenture.com}
}

\begin{document}
\maketitle

\begin{abstract}
Automatic AV-sync metrics are widely used to rank and train audio-visual generators, but they are rarely
audited as measurement instruments. We jointly audit AV-Align, ImageBind AV-relevance, JavisScore, and
Synchformer/DeSync under a common reliability protocol: controlled-distortion monotonicity, preprocessing
sensitivity, rank uncertainty, cross-metric agreement, PEAVS-proxy agreement, and learned fusion. The result is
an \emph{axis split}, not a single winner: Synchformer/DeSync is the strongest temporal-offset tracker
($\tau=0.84$), ImageBind/JavisScore better match the PEAVS human-aligned proxy ($\tau=0.20$) and
content-disruption families, and AV-Align is the weakest standalone metric. The metrics mutually disagree
(Krippendorff $\alpha=0.066$), and neither linear nor simple $k$-NN fusion improves PEAVS agreement over the
best individual metric. We recommend reporting AV-sync as a \emph{Reliability Card} (metric-family breakdowns
with confidence intervals) rather than a single bare synchronization score.
\end{abstract}

\section{Introduction}
Audio-visual (AV) generation, spanning joint text-to-AV, video-to-audio (Foley), and audio-to-video,
is advancing on the strength of automatic \emph{synchronization} metrics. Benchmarks rank
models by a single sync number, and preference-optimization pipelines now \emph{train} on a sync
signal, so an unreliable metric can corrupt optimization itself. The dominant metrics are AV-Align~\cite{avalign} (optical-flow motion peaks vs.\
audio-onset peaks, IoU), Synchformer/DeSync~\cite{synchformer} (a learned offset predictor),
JavisScore~\cite{javisdit} (windowed ImageBind AV similarity), and ImageBind
AV-relevance~\cite{imagebind} (semantic cosine). JavisScore was introduced \emph{because}
AV-Align ``may produce misleading results'' on complex scenes, so the community already
suspects metric-specific failure modes, yet the deployed metric family has not been jointly audited under a common reliability protocol.

A synchronization score is a \emph{measuring instrument}: it should respond monotonically to known
errors, stay stable under benign preprocessing, agree with related instruments, and align with perceptual
judgment. Yet the deployed AV-sync instruments have never been jointly audited on these properties.
Synthetic controlled-degradation meta-evaluation, testing whether a metric tracks \emph{known}
quality changes, is an established paradigm (SLVMEval~\cite{slvmeval} for text-to-long-video
quality; STREAM~\cite{stream} for video corner-cases), but it has never been applied to
\emph{audio-visual synchronization} metrics, nor combined with cross-metric and stability
analysis. Prior AV work either proposes a \emph{new} human-aligned metric and validates only
itself (PEAVS~\cite{peavs}; AVBench~\cite{avbench}) or \emph{uses} the metrics to rank models
(VABench~\cite{vabench}). The nearest precedent, JavisDiT~\cite{javisdit}, releases a
3{,}000-sample human set and shows JavisScore is more \emph{accurate} than AV-Align, validating
one proposed metric, not auditing the deployed set for stability, cross-metric agreement, or
rank-flip. Our novelty is thus one of \emph{domain and scope}: to our knowledge, the first joint reliability audit
of the deployed AV-sync metric family under a common protocol.

We close that gap. Our contributions: \textbf{(1)} a reproducible, annotation-free
\textbf{synthetic-desynchronization oracle} imposing known desync (temporal shift, audio
speed, fragment shuffle, intermittent mute) on real clips; \textbf{(2)} a \textbf{preprocessing-sensitivity
harness} quantifying each metric's coefficient of variation under should-not-change resampling
and its rank-flip probability; \textbf{(3)} the first \textbf{cross-metric agreement}
measurement across the deployed metrics, plus a PEAVS-proxy comparison; and \textbf{(4)} a
\textbf{learned, conformal-calibrated meta-metric} evaluated against the metrics as baselines,
testing (and finding wanting) the hypothesis that they combine into a human-aligned score. The
headline: competence is \emph{axis-specific}; Synchformer tracks temporal offset far better than the rest
yet is weakly PEAVS-aligned, the metrics mutually disagree ($\alpha\!=\!0.066$), and none wins both axes.

\section{Related Work}
\noindent\textbf{AV-synchronization metrics.} Automatic sync scoring spans two families. \emph{Signal-level}
metrics compare low-level motion and audio events: AV-Align~\cite{avalign} matches optical-flow motion peaks
to audio-onset peaks by IoU. \emph{Learned} metrics embed both modalities: Synchformer/DeSync~\cite{synchformer}
is a transformer trained to predict the audio-visual temporal offset; ImageBind~\cite{imagebind} yields a
generic cross-modal cosine; and JavisScore~\cite{javisdit} aggregates windowed ImageBind similarity, introduced
explicitly because AV-Align ``may produce misleading results.'' Both target \emph{semantic} correspondence by
design, not onset-level timing (ImageBind's training negatives swap the entire paired audio and encode only two
frames), so weak temporal-oracle tracking is expected for them. PEAVS~\cite{peavs} instead trains a
\emph{human-aligned} predictor on 120K opinion scores. These metrics are deployed as leaderboard criteria, yet
each was validated (if at all) only against its own design goal; none has been audited for stability,
cross-metric agreement, or human grounding as a group. We treat all four signal/learned metrics as the
black boxes the community actually uses and ask whether their \emph{scores} can be trusted.

\noindent\textbf{Meta-evaluation by controlled degradation.} Testing whether a metric tracks \emph{known},
synthetically-imposed quality changes is an established meta-evaluation paradigm: SLVMEval~\cite{slvmeval}
probes text-to-long-video quality metrics, and STREAM~\cite{stream} stress-tests video metrics on temporal
corner cases. This annotation-free oracle has not been applied to audio-visual \emph{synchronization}, nor
combined with preprocessing sensitivity and inter-metric agreement; we port it to AV-sync and add a PEAVS-proxy axis.

\noindent\textbf{AV generation and benchmarks.} Video-to-audio and joint AV generation
(MMAudio~\cite{mmaudio}, ASVA~\cite{asva}, JavisDiT~\cite{javisdit}) are ranked by the very sync metrics we
audit; benchmarks (VABench~\cite{vabench}, AVBench~\cite{avbench}) aggregate them but validate at most one
proposed metric, not the deployed set. Our audit is orthogonal, and adds a conformal-calibrated~\cite{conformal}
meta-metric tested as a would-be combiner.

\section{Method}
We score each metric as a black box, $f(\text{clip})\in\mathbb{R}$ (higher $=$ better synced).

\noindent\textbf{Synthetic-desync oracle (agreement without humans).} On a pool of real,
well-synced clips we apply a controlled desync of known magnitude $m$ from four families
(temporal shift, audio speed, fragment shuffle, intermittent mute), each an increasing-desync
grid, so the ground-truth ordering is known by construction. The oracle is a monotonicity
test: as \emph{added} distortion grows the score should fall, regardless of the clip's baseline
sync. Per clip and family we compute Kendall $\tau$ between the metric score and \emph{negative}
perturbation severity (i.e.\ the ideal ``more-distortion-is-worse'' order), so that
$\tau\!=\!1$ denotes perfectly monotonic degradation as distortion increases and $\tau\!=\!0$
denotes no response; a faithful metric yields $\tau\!\to\!1$. We report per-family mean $\tau$ with clip-bootstrap
95\% CIs. The four families are PEAVS~\cite{peavs}-inspired synchrony-related distortions; only temporal shift
is a pure global AV offset, while speed, shuffle, and mute probe broader temporal-structure and content-disruption sensitivity.

\noindent\textbf{Preprocessing-sensitivity harness.} A metric should be invariant to changes that do not alter
true sync. We resample clips (random fixed-length crops; length truncation) and report the
coefficient of variation (CV). We further compute the \textbf{rank-flip probability}: under
clip-bootstrap, how often a metric fails to order two conditions that differ by one known
desync step, a direct test of whether single-number gaps of that size are within noise.

\noindent\textbf{Cross-metric agreement.} Treating each metric as a ``rater'' of clip sync,
we compute pairwise Kendall $\tau$ (rank-based) and Krippendorff $\alpha$ (interval) on per-clip scores that
are first \emph{z-scored per metric}, so agreement is scale-free and unaffected by the metrics' differing
ranges; we use PEAVS's human inter-annotator $\alpha\!\approx\!0.71$ as an anchor.

\noindent\textbf{Learned calibrated meta-metric (method).} We test whether the metrics can be
\emph{combined} into a human-aligned score: a regularized linear model (ridge) regresses the
human-aligned PEAVS score on the per-clip metric scores, evaluated strictly out-of-fold (5-fold
and leave-one-out, no leakage), with split-conformal prediction intervals for calibrated
per-clip uncertainty. We compare its held-out Kendall $\tau$ vs.\ PEAVS to every single metric.

\section{Experiments}\label{sec:exp}
\textbf{Clips.} AVSync15~\cite{asva} (15 classes, 1{,}500 real highly-synced clips from
VGGSound); the primary audit uses 75 clips (5/class), the subset on which PEAVS scoring and all auxiliary
analyses (crop sensitivity, generated-audio pairing, qualitative inspection) are aligned; a $2\times$
replication (150 clips) reproduces every conclusion (Synchformer temporal $\tau\!=\!0.87$, $\alpha\!=\!0.07$; \S\ref{sec:limits}). \textbf{Metrics.} We reimplement AV-Align from the published
TempoTokens~\cite{avalign} algorithm (optical-flow motion peaks vs.\ audio-onset peaks matched by IoU) with
its default peak-picking parameters (validated against the official code, App.~B); ImageBind AV-relevance and JavisScore use the official ImageBind encoder
(JavisScore: 2\,s windows, 1.5\,s overlap, mean of the 40\% least-synced windows). For Synchformer/DeSync we
use the released checkpoint and score each clip as $-|\mathbb{E}[\Delta]|$, the negative absolute \emph{expected}
audio-visual offset under the softmax over the model's offset-class grid (App.~A: the split is unchanged under $-\mathbb{E}[|\Delta|]$ and the modal offset); inputs are reencoded to the released
25\,fps / $256$px / 16\,kHz spec and loop-padded to the fixed 5\,s window. Every other metric uses its native
preprocessing; scores are oriented so higher $=$ better synchronization. \textbf{Stats.}
Kendall $\tau$-b, $\ge$1k-resample clip-bootstrap 95\% CIs, preprocessing-sensitivity CV, bootstrap rank-flip.
All reproducible from committed code, seeds, and SLURM scripts.

\begin{table*}[t]\centering\small
\begin{tabular}{l cccc c cc c}
\toprule
& \multicolumn{4}{c}{Oracle tracking, Kendall $\tau$ $\uparrow$} & Preproc. & \multicolumn{2}{c}{Rank-flip $\downarrow$} & PEAVS \\
\cmidrule(lr){2-5}\cmidrule(lr){7-8}
Metric & Shift & Speed & Shuffle & Mute & CV $\downarrow$ & Shift & Worst-fam. & proxy $\tau$ $\uparrow$ \\
\midrule
AV-Align~\cite{avalign} (reimpl.)   & 0.21 & 0.04 & 0.01 & 0.24 & 0.54 & 0.66 & 0.96 & 0.00 \\
ImageBind-rel.~\cite{imagebind}     & 0.16 & 0.37 & \textbf{0.38} & 0.61 & 0.15 & 0.54 & \textbf{0.69} & \textbf{0.20} \\
JavisScore~\cite{javisdit}          & 0.39 & 0.51 & 0.31 & \textbf{0.73} & \textbf{0.14} & 0.46 & 0.90 & \textbf{0.20} \\
Synchformer/DeSync~\cite{synchformer} & \textbf{0.84} & \textbf{0.76} & 0.27 & 0.59 & n/a & \textbf{0.19} & 1.00 & 0.07 \\
\bottomrule
\end{tabular}
\caption{\textbf{Main reliability scorecard} on AVSync15 (best per column \textbf{bold}; higher $\tau$, lower
CV/flip better). Synchformer/DeSync leads temporal tracking; ImageBind/JavisScore lead PEAVS-proxy agreement (a
tie); AV-Align is the weakest standalone metric. \textbf{No metric wins both the temporal-oracle and
PEAVS-proxy axes.} Rank-flip \emph{Shift}/\emph{Worst-fam.}\ $=$ pure-shift vs.\ maximum-family adjacent
mis-ordering; Synchformer CV n/a (fixed $\ge5$\,s input). The AV-Align row is our reimplementation; official
TempoTokens scoring gives the same aggregate conclusion (App.~B). CIs are in the supplement.}
\label{tab:master}
\end{table*}

\begin{figure}[t]\centering
\includegraphics[width=0.75\linewidth]{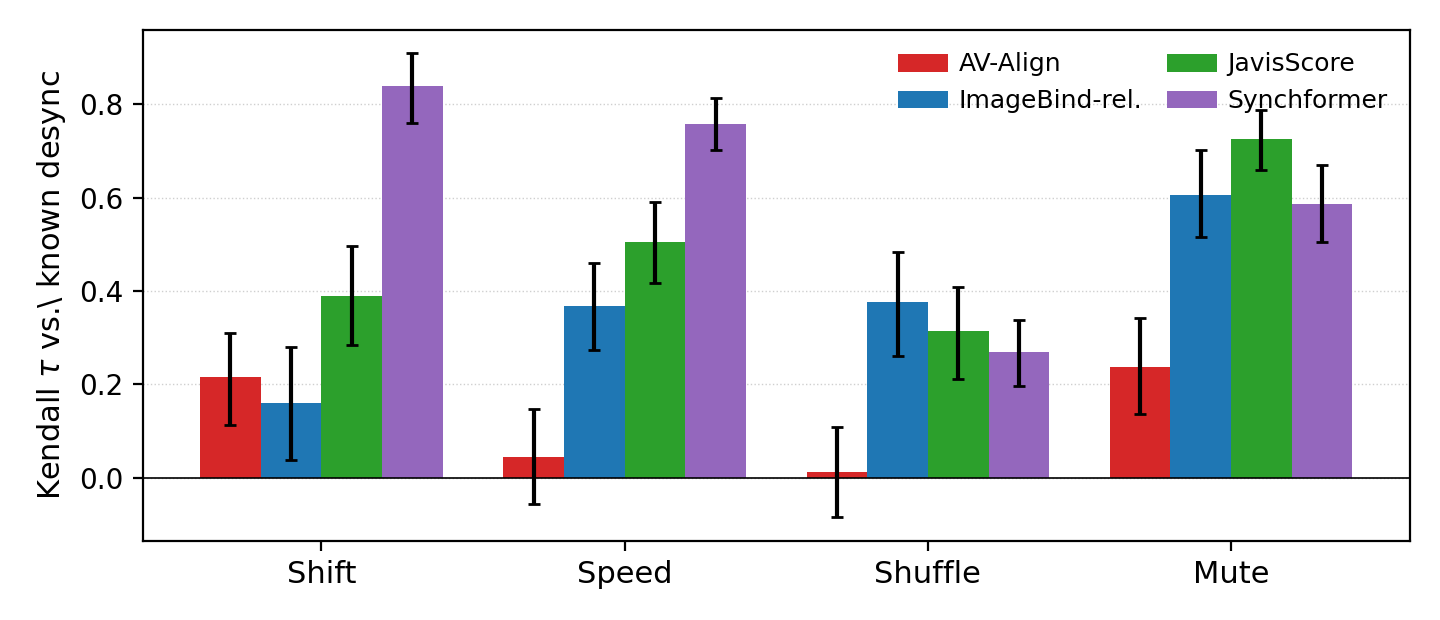}
\caption{Agreement with controlled desynchronization (Kendall $\tau$, 95\% CI).
\textbf{Synchformer leads} temporal and audio-speed tracking, while JavisScore/ImageBind respond
more strongly to intermittent mute; AV-Align is weakest overall (CI crosses zero on audio-speed).}
\label{fig:oracle}
\end{figure}

\section{Results}
Table~\ref{tab:master} consolidates the full audit. The main conclusion is an \emph{axis split}, not a
single winning metric; we walk through it below.

\subsection{Metrics split by reliability axis}
\noindent\textbf{Oracle tracking (Fig.~\ref{fig:oracle}).} The dedicated learned offset predictor
\textbf{Synchformer tracks the pure temporal-offset axes far better than the rest}: temporal-shift $\tau=0.84$
$[0.78,0.90]$ and audio-speed $0.76$, versus $0.16$--$0.39$ (temporal) for the others. But it does
\emph{not} generalize to content disruption: on shuffle ImageBind leads ($0.38$ vs.\ Synchformer's
$0.27$) and on mute JavisScore leads ($0.73$ vs.\ $0.59$). So metric competence is \emph{axis-specific}: the
offset predictor owns temporal ordering, the embedding metrics own content disruption, and AV-Align is the
weakest oracle tracker overall (near chance on audio-speed, $0.04$, and shuffle, $0.01$).

\subsection{Small leaderboard gaps are not always resolvable}
\noindent\textbf{Preprocessing sensitivity and rank-flip.} AV-Align is the \emph{only} high-variance metric
(median CV $0.54$ vs.\ $\le0.15$). The \emph{worst-family} adjacent flip (bootstrap probability of mis-ordering
two adjacent desync levels, i.e.\ consecutive grid magnitudes; max over non-identity pairs) is high for every metric: $0.96$/$0.69$/$0.90$/$1.00$
(AV-Align/ImageBind/JavisScore/Synchformer), but this is each metric's weak family. On the
\emph{temporal-shift} ranking that matters most, the flip is far lower for Synchformer ($0.19$, and $0.00$ on
speed) than for the others ($0.46$--$0.66$); its $1.00$ is shuffle alone. So Synchformer is reliable
for temporal ranking even though no metric is reliable across all families. Adjacent-level discriminability
($d'\!=\!\Delta\mu/\sigma_\text{clip}$, the mean score gap between adjacent levels over clip noise) agrees: only
Synchformer clears $d'\!\ge\!1$, on temporal shift ($d'\!=\!1.6$); elsewhere $d'\!<\!1$ (adjacent levels overlap
$>16\%$). Leaderboard gaps below a metric's minimum detectable difference (MDD, the smallest gap whose bootstrap
CI excludes zero; $13$--$17\%$ of range) are noise.

\noindent\textbf{Real generations: close models break the similarity metrics.} We generate audio
for 45 AVSync15 videos with three MMAudio~\cite{mmaudio} checkpoints and ask whether each metric
reliably ranks them (paired clip-bootstrap flip probability, Table~\ref{tab:mmaudio}). A \emph{far} pair
(small-16k vs.\ large-44k) is easy for all metrics, but for a \emph{close} pair (large-44k vs.\ large-44k-v2),
the leaderboard regime of similar systems, the similarity metrics' rankings fall within noise while
\textbf{only Synchformer reliably separates the generators}. Re-running the
oracle on the 45 generated clips reproduces the ranking (Synchformer best, $\tau\!=\!0.76$; AV-Align worst),
so the conclusions are not an artifact of natural clips. This is not a full MMAudio benchmark but a
leaderboard-resolution stress test: paired bootstrap uncertainty on matched generated outputs.

\begin{table}[H]\centering\small
\begin{tabular}{lcc}
\toprule
Metric & Far model gap & Close model gap \\
\midrule
AV-Align & $\le0.09$ & $0.35$ \\
ImageBind-rel. & $\le0.09$ & $0.33$ \\
JavisScore & $\le0.09$ & $0.35$ \\
Synchformer/DeSync & $\le0.09$ & $\mathbf{0.08}$ \\
\bottomrule
\end{tabular}
\caption{\textbf{Leaderboard risk on generated outputs}: paired clip-bootstrap flip probability~$\downarrow$.
Far model gaps are robust across metrics; a close pair falls within similarity-metric noise, and only
Synchformer separates it.}\label{tab:mmaudio}
\end{table}

\subsection{No single score or simple fusion solves the problem}
\begin{figure}[t]\centering
\includegraphics[width=0.44\linewidth]{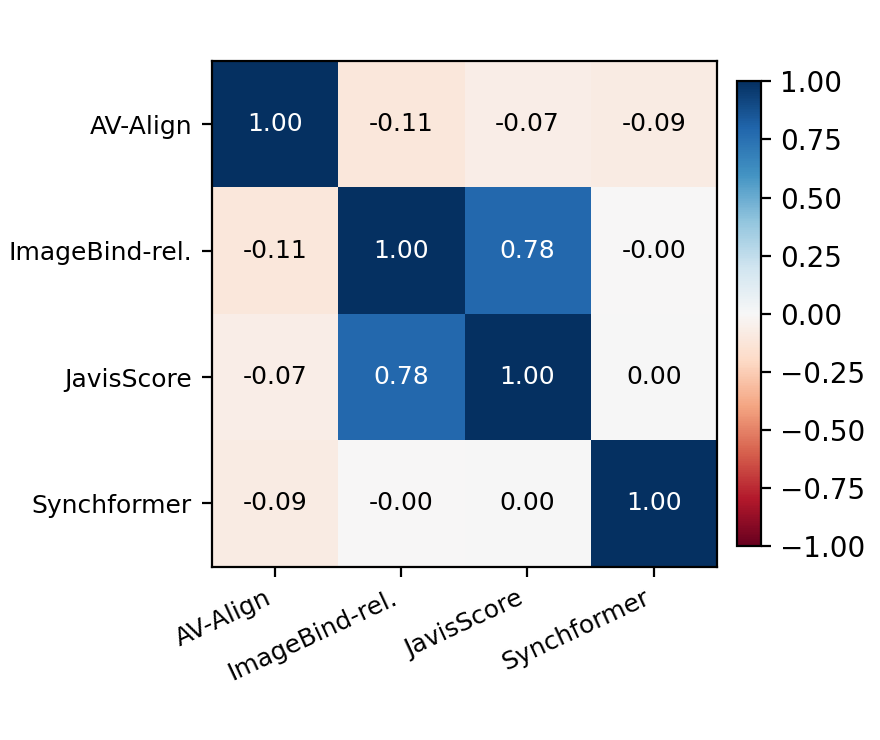}\hfill
\includegraphics[width=0.44\linewidth]{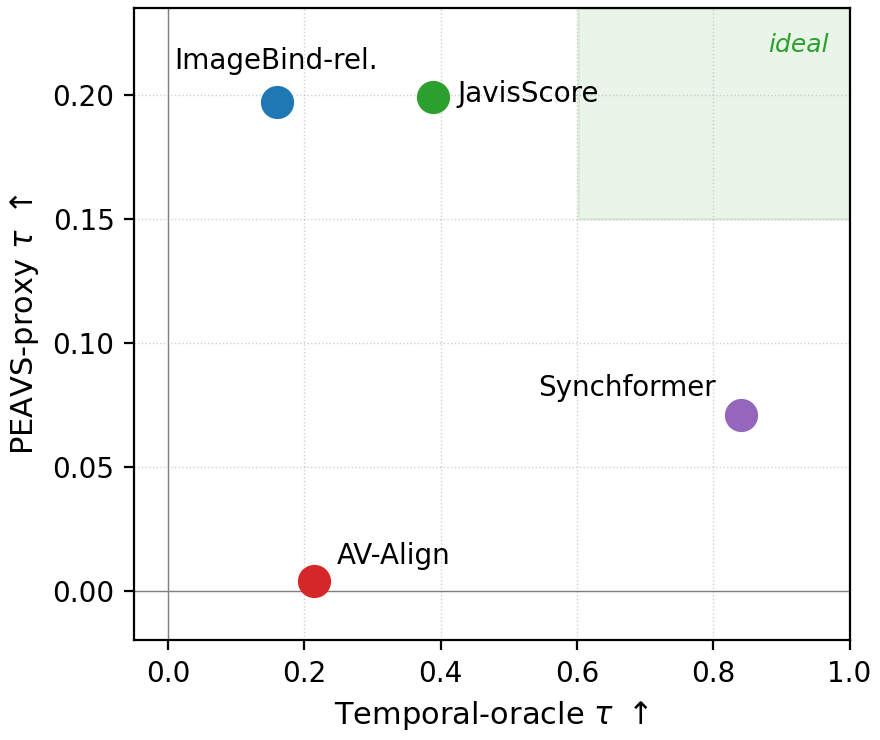}
\caption{Left: cross-metric agreement ($\alpha=0.066$); only the shared-backbone embedding pair
agrees. Right: the two axes, oracle tracking (x) vs.\ PEAVS agreement (y), are nearly
orthogonal; no metric is high on both.}
\label{fig:cross}\label{fig:axes}
\end{figure}

\noindent\textbf{Cross-metric disagreement (Fig.~\ref{fig:cross}, left).} Treating each metric as
a rater, the four agree almost not at all: Krippendorff $\alpha=0.066$ ($z$-scored; rises to $0.21$ on the
controlled grid, App.~C), far below the inter-annotator agreement reported in PEAVS ($\approx0.71$). Every pairwise Kendall $\tau$ is near zero except
ImageBind-rel.\ vs.\ JavisScore ($+0.78$), which is expected and \emph{not} independent (shared
ImageBind backbone). Crucially, Synchformer agrees with the embedding metrics and AV-Align at $\tau\approx0$: an
oracle-accurate metric orders real clips unlike any deployed one. A PCA confirms the four metrics span
\emph{three} orthogonal axes (only the shared-backbone ImageBind/JavisScore pair duplicates, $r\!=\!0.93$): a
single sync number projects three dimensions onto one. On already-synced clips this disagreement is over
synchronization \emph{evidence}, not \emph{quality}; it persists under the oracle and on generated clips, so it
is not a clean-clip range artifact.

\noindent\textbf{Two distinct axes: oracle vs.\ PEAVS (Fig.~\ref{fig:axes}).} Scoring the same
clips with PEAVS~\cite{peavs}, a human-aligned automatic proxy reported to correlate with human ratings,
Synchformer (best at the oracle) agrees with PEAVS at only
$\tau=0.07$, while ImageBind/JavisScore tie for the highest PEAVS-proxy agreement ($\tau=0.20$) yet only moderately track
the oracle; AV-Align is $\approx0$ on both. \textbf{No metric is good at both, and the two axes
are nearly orthogonal}: detecting controlled desynchronization and matching the PEAVS-aligned
perceptual proxy are different problems. Scene-complexity stratification (clips split at the median optical-flow
magnitude, equal halves) is diagnostic: AV-Align is
weaker on low-motion than high-motion clips ($\tau=0.18$ vs.\ $0.25$), consistent with its reliance on motion
peaks, while JavisScore's PEAVS agreement is much higher in high-motion than low-motion scenes ($\tau=0.48$
vs.\ $0.04$); Synchformer's temporal lead is scene-invariant.

\noindent\textbf{A learned meta-metric cannot fix it.} A ridge meta-metric over the four scores,
evaluated strictly out-of-fold with split-conformal intervals, never beats the best single metric (held-out $\tau$ vs.\ PEAVS $\le0.12$ vs.\ $0.20$). This is
not a limitation of \emph{linear} fusion: a simple nonlinear combiner (leave-one-out $k$-NN)
does no better ($\le0$ vs.\ PEAVS). In our sample, then, neither linear nor simple nonlinear fusion improves
PEAVS agreement over the best single metric: naive fusion is insufficient here, though it does not rule out
supervised calibration with direct human labels.

\section{Discussion and Guidance}\label{sec:limits}
We audit on AVSync15; the axis-specific split also replicates on a second, in-the-wild domain (VGGSound,
App.~G). The temporal oracle uses audio-lag shifts; a bidirectional variant (audio lead/lag, released
harness) preserves the ranking. Our
perceptual grounding uses the PEAVS \emph{metric} (human-aligned, Pearson $0.79$), a scalable proxy, not fresh
human labels; PEAVS is itself a learned model and may share representational biases with the embedding metrics,
so we read this axis as \emph{PEAVS agreement}, not perceptual ground truth. Direct human preference annotation
remains future work before any claim of perceptual superiority. The findings are not sample-size artifacts: the 150-clip replication (\S\ref{sec:exp}) reproduces
every structural conclusion. The oracle imposes \emph{controlled} distortions, not generative artifacts
(complemented by the MMAudio audit above).

\begin{table}[H]\centering\small
\begin{tabular}{ll}
\toprule
Reliability axis & Report \\
\midrule
Temporal oracle & $\tau$ + CI \\
Content/distortion oracle & $\tau$ + CI \\
Preprocessing sensitivity & CV + CI \\
Ranking resolution & flip prob. \\
PEAVS proxy & $\tau$ + CI \\
Cross-metric consistency & $\tau$ / $\alpha$ \\
Direct human calibration & pairwise acc.\ (if avail.) \\
\bottomrule
\end{tabular}
\caption{\textbf{Reliability Card}: report AV-sync on these axes, not one bare number (full version, App.~E).}\label{tab:cardmain}
\end{table}

\noindent\textbf{Guidance.} The AV-generation community ranks (and increasingly trains) models on
synchronization metrics that mutually disagree ($\alpha=0.066$ vs.\ $\approx0.71$ among PEAVS's human
annotators). Do not report AV-Align alone: the weakest, most preprocessing-sensitive standalone metric; for
\emph{temporal} synchronization prefer Synchformer/DeSync ($\tau=0.84$), but its competence is
\emph{axis-specific}: it fails on content-disruption families. No metric wins both axes, and no meta-metric
combines them beyond the best single one. The field should stop reporting AV-sync as a single bare number;
until directly human-calibrated metrics exist, authors should report a \emph{Reliability
Card}~(Table~\ref{tab:cardmain}): metric-family scores, uncertainty, and rank resolution.

{\small\bibliographystyle{ieeenat_fullname}\bibliography{refs}}

\begin{thebibliography}{12}
\providecommand{\natexlab}[1]{#1}
\providecommand{\url}[1]{\texttt{#1}}
\expandafter\ifx\csname urlstyle\endcsname\relax
  \providecommand{\doi}[1]{doi: #1}\else
  \providecommand{\doi}{doi: \begingroup \urlstyle{rm}\Url}\fi

\bibitem[Cheng et~al.(2025)]{mmaudio}
Ho~Kei Cheng et~al.
\newblock {MMAudio}: Taming multimodal joint training for high-quality video-to-audio synthesis.
\newblock \emph{CVPR}, 2025.
\newblock arXiv:2412.15322.

\bibitem[Girdhar et~al.(2023)]{imagebind}
Rohit Girdhar et~al.
\newblock {ImageBind}: One embedding space to bind them all.
\newblock \emph{CVPR}, 2023.

\bibitem[Goncalves et~al.(2024)]{peavs}
Lucas Goncalves et~al.
\newblock {PEAVS}: Perceptual evaluation of audio-visual synchrony grounded in viewers' opinion scores.
\newblock \emph{ECCV}, 2024.
\newblock arXiv:2404.07336.

\bibitem[Hua et~al.(2026)]{vabench}
Daili Hua et~al.
\newblock {VABench}: A comprehensive benchmark for audio-video generation.
\newblock \emph{CVPR}, 2026.
\newblock arXiv:2512.09299.

\bibitem[Iashin et~al.(2024)]{synchformer}
Vladimir Iashin et~al.
\newblock {Synchformer}: Efficient synchronization from sparse cues.
\newblock \emph{ICASSP}, 2024.
\newblock arXiv:2401.16423.

\bibitem[Kim et~al.(2024)]{stream}
Pum~Jun Kim et~al.
\newblock {STREAM}: Spatio-temporal evaluation and analysis metric for video generative models.
\newblock \emph{arXiv:2403.09669}, 2024.

\bibitem[Liu et~al.(2025)]{javisdit}
Kai Liu et~al.
\newblock {JavisDiT}: Joint audio-video diffusion transformer with hierarchical spatio-temporal prior synchronization.
\newblock \emph{arXiv:2503.23377}, 2025.

\bibitem[Matsuda et~al.(2026)]{slvmeval}
Ryosuke Matsuda et~al.
\newblock {SLVMEval}: Synthetic meta-evaluation benchmark for text-to-long video generation.
\newblock In \emph{CVPR}, 2026.
\newblock arXiv:2603.29186.

\bibitem[Shafer and Vovk(2008)]{conformal}
Glenn Shafer and Vladimir Vovk.
\newblock A tutorial on conformal prediction.
\newblock \emph{JMLR}, 2008.

\bibitem[Yang et~al.(2026)]{avbench}
Jialiang Yang et~al.
\newblock {AVBench}: Human-aligned and automated evaluation benchmark for audio-video generative models.
\newblock \emph{arXiv:2605.24652}, 2026.

\bibitem[Yariv et~al.(2024)]{avalign}
Guy Yariv et~al.
\newblock Diverse and aligned audio-to-video generation via text-to-video model adaptation.
\newblock \emph{AAAI}, 2024.
\newblock arXiv:2309.16429.

\bibitem[Zhang et~al.(2024)]{asva}
Lin Zhang et~al.
\newblock Audio-synchronized visual animation.
\newblock \emph{ECCV}, 2024.
\newblock arXiv:2403.05659.

\end{thebibliography}

\clearpage
\onecolumn
\appendix
\section*{Appendix (supplementary material)}
\small

\noindent\textbf{A. Synchformer scoring-reduction ablation.} The main text reduces the predicted
offset posterior to $-|\mathbb{E}[\Delta]|$, which can cancel for a bimodal posterior. Two
non-cancelling reductions, $-\mathbb{E}[|\Delta|]$ (expected absolute offset) and $-|\Delta_{\arg\max}|$ (modal offset), leave the temporal/
perceptual split intact (Table~\ref{tab:sfabl}): temporal-oracle $\tau$ stays high and PEAVS $\tau$
stays low under all three, so Synchformer's conclusions do not depend on the reduction.

\begin{table}[H]\centering
\begin{tabular}{lcc}
\toprule
Synchformer reduction & Temporal-shift $\tau$ & PEAVS $\tau$ \\
\midrule
$-|\mathbb{E}[\Delta]|$ (main) & $0.84$ & $0.07$ \\
$-\mathbb{E}[|\Delta|]$        & $0.81$ & $0.10$ \\
$-|\Delta_{\arg\max}|$ (modal)  & $0.78$ & $-0.05$ \\
\bottomrule
\end{tabular}
\caption{The temporal/perceptual split survives the score reduction ($75$ clips).}\label{tab:sfabl}
\end{table}

\noindent\textbf{B. AV-Align: official vs.\ our implementation.} We ran the official TempoTokens AV-Align
through the same oracle (Table~\ref{tab:avalign}). Every \emph{aggregate} conclusion holds under the official
code: both versions are weak oracle trackers (temporal $\tau\approx0.22$; near-zero on speed/fragment), highly
variable (CV $0.34$--$0.54$), unreliable for temporal ranking (temporal flip $\approx0.67$), and near-zero
PEAVS. Per-clip scores correlate only weakly across the two (Kendall $\tau=0.11$, $n=45$), reflecting
AV-Align's sensitivity; because the aggregate audit agrees, the main table reports our reimplementation and the
AV-Align conclusions are robust to implementation, not a reimplementation artifact.

\begin{table}[H]\centering
\begin{tabular}{lcccccc}
\toprule
AV-Align & Temp. & Speed & Frag. & Mute & CV & PEAVS \\
\midrule
Official (TempoTokens) & $0.23$ & $0.10$ & $-0.10$ & $0.35$ & $0.34$ & $-0.04$ \\
Ours (main table)      & $0.21$ & $0.04$ & $0.01$  & $0.24$ & $0.54$ & $0.00$ \\
\bottomrule
\end{tabular}
\caption{Official vs.\ reimplemented AV-Align through the oracle ($75$ clips): the aggregate pattern (weak
tracking, high variance, near-zero PEAVS) is the same. Temporal rank-flip is $0.68$ (official) vs.\ $0.66$ (ours).}\label{tab:avalign}
\end{table}

\noindent\textbf{C. Cross-metric agreement across settings.} All scores are $z$-scored per metric
(scale-free) before Krippendorff $\alpha$. Agreement rises with the strength of the sync signal
(Table~\ref{tab:alpha}): near zero on clean clips, $3\times$ higher on the controlled grid where sync
varies by design. The low clean-clip $\alpha$ thus reflects subtle clean-clip variation, not
uninterpretable disagreement.

\begin{table}[H]\centering
\begin{tabular}{lcc}
\toprule
Setting & Krippendorff $\alpha$ & $n$ \\
\midrule
Clean clips             & $0.07$ & $75$ \\
Controlled desync grid  & $0.21$ & $1650$ \\
Generated (MMAudio)     & $0.11$ & $45$ \\
\bottomrule
\end{tabular}
\caption{Cross-metric $\alpha$ by setting ($z$-scored).}\label{tab:alpha}
\end{table}

\noindent\textbf{D. Preprocessing-sensitivity measure.} We report CV over should-not-change resampling
on each metric's positive score range, complemented by the scale-free rank-flip probability; the two
agree (AV-Align is the outlier on both), so the CV comparison is not an artifact of differing score ranges.

\clearpage
\noindent\textbf{E. Metric Reliability Card.} We distill the audit into a reusable reporting standard
(Table~\ref{tab:card}): an AV-sync metric should be characterized on all seven axes, not by a single number. We recommend
authors report this card for any metric they use to rank or train models.

\begin{table}[H]\centering
\begin{tabular}{lll}
\toprule
Reliability axis & What it tests & Report \\
\midrule
Temporal oracle       & global-offset sensitivity      & $\tau$ + CI \\
Distortion oracle     & speed/shuffle/mute sensitivity & $\tau$ + CI \\
Preproc.\ sensitivity & crop/length invariance         & CV + CI \\
Ranking resolution    & leaderboard uncertainty        & flip prob. \\
PEAVS proxy           & perceptual alignment           & $\tau$ + CI \\
Cross-metric consist. & agreement with peers           & rank $\tau$/$\alpha$ \\
Direct human calibration & direct preference check    & pairwise acc. \\
\bottomrule
\end{tabular}
\caption{The Metric Reliability Card: the axes any deployed AV-sync metric should be reported on.}\label{tab:card}
\end{table}

\vspace{2pt}
\noindent\textbf{F. Qualitative disagreement examples.} Concrete AVSync15 clips where the metrics diverge
(per-metric $z$-scores over the audit set): \emph{(i)}~a clip the embedding metrics rate well-synced
(ImageBind/JavisScore $z\!\approx\!+1.0$) but PEAVS rates poorly ($z\!=\!-2.2$), semantic AV-relevance
mistaken for synchrony; \emph{(ii)}~a clip Synchformer flags as badly offset ($z\!=\!-6.0$) while AV-Align,
ImageBind, JavisScore \emph{and} PEAVS all rate it synced ($z\!\ge\!+0.8$), the offset predictor firing where
perception sees none; \emph{(iii)}~a clip AV-Align rates well-synced ($z\!=\!+1.4$) while the other three
metrics and PEAVS disagree ($z\!\le\!-0.6$), a spurious optical-flow/onset coincidence. These make concrete
the axis-specific, mutually-inconsistent behavior quantified in the main text.

\vspace{2pt}
\noindent\textbf{G. Second-domain replication (VGGSound).} We rerun the oracle on $75$ in-the-wild VGGSound
clips (unconstrained YouTube AV, distinct from curated AVSync15). The axis-specific split replicates
(Table~\ref{tab:vgg}): Synchformer is by far the strongest temporal tracker ($\tau=0.63$ vs.\ $\le0.27$ for
the others), the embedding metrics lead on content disruption (ImageBind/JavisScore up to $0.36/0.59$ on
fragment and mute), and AV-Align is weakest overall (near-zero on speed and fragment). Absolute $\tau$ is
lower than on curated AVSync15 (expected for noisier in-the-wild clips) but the \emph{ordering}, the
paper's headline, holds.

\begin{table}[H]\centering
\begin{tabular}{lcccc}
\toprule
Metric (VGGSound) & Temporal & Aud.-speed & Fragment & Mute \\
\midrule
AV-Align        & $0.16$ & $0.04$ & $-0.03$ & $0.24$ \\
ImageBind-rel.  & $0.04$ & $0.26$ & $\mathbf{0.36}$ & $0.46$ \\
JavisScore      & $0.27$ & $0.36$ & $0.25$ & $\mathbf{0.59}$ \\
Synchformer     & $\mathbf{0.63}$ & $\mathbf{0.55}$ & $0.17$ & $0.33$ \\
\bottomrule
\end{tabular}
\caption{Oracle Kendall $\tau$ on VGGSound: the temporal (Synchformer) / content (embedding) split and
AV-Align's weakness replicate on a second domain.}\label{tab:vgg}
\end{table}

\end{document}